%% file: bmvc_final.tex
\documentclass{bmvc2k}

\usepackage{diagbox}
\usepackage{amssymb}
\usepackage{pifont}
\usepackage{multirow}
\usepackage{amsfonts}
\usepackage{amsmath}
\usepackage{color}
\usepackage{makecell}
\usepackage{rotating}
\newcommand\Xuanlong{}
\newcommand\Emi{}
\DeclareMathOperator*{\argmin}{argmin}
\title{SLURP: Side Learning Uncertainty for Regression Problems}

\addauthor{Xuanlong Yu}{xuanlong.yu@universite-paris-saclay.fr}{1}
\addauthor{Gianni Franchi}{gianni.franchi@ensta-paris.fr}{2}
\addauthor{Emanuel Aldea}{emanuel.aldea@universite-paris-saclay.fr}{1}

\addinstitution{
 SATIE, Paris-Saclay University\\
 Gif-sur-Yvette, France
}
\addinstitution{
 U2IS, ENSTA Paris\\
 Institut Polytechnique de Paris\\
 Palaiseau, France
}

\runninghead{Yu et al.}{Side Learning Uncertainty for Regression Problems with SLURP}

\def\eg{\emph{e.g}\bmvaOneDot}
\def\Eg{\emph{E.g}\bmvaOneDot}
\def\etal{\emph{et al}\bmvaOneDot}

\begin{document}

\maketitle

\begin{abstract}
It has become critical for deep learning algorithms to quantify their output uncertainties to satisfy reliability constraints and provide accurate results.
Uncertainty estimation for regression has received less attention than classification due to the more straightforward standardized output of the latter class of tasks and their high importance. However, regression problems are encountered in a wide range of applications in computer vision.  We propose SLURP, a generic approach for regression uncertainty estimation via a side learner that exploits the output and the intermediate representations generated by the main task model. We test SLURP on two critical regression tasks in computer vision:  monocular depth and optical flow estimation. In addition, we conduct exhaustive benchmarks comprising transfer to different datasets and the addition of aleatoric noise. The results show that our proposal is generic and readily applicable to various regression problems and has a low computational cost with respect to existing solutions.
\end{abstract}

\input{intro}

\input{related}

\section{Our method}
\subsection{Problem description and motivation}
\label{3.1}
Our training procedure is a two-step process that first consists of training a DNN $f$ to perform its main regression task. Then in a second stage, the parameters in $f$ are fixed and will not be updated any more. Based on a primary DNN $f$, we train a second DNN $g$ to predict the uncertainty of the first DNN.

\textbf{Task 1}:  given dataset $D=\{(\mathbf{x}_i, y_i)\}_i$, we write 
$P(Y|\mathbf{x})$  the conditional distribution for the ground truth value given an input value $\mathbf{x}$ (\emph{e.g} an RGB image).
Let us denote $f$, the main task predictor, which is trained by minimizing the objective function $l_f(f(\mathbf{x}), y)$ over the dataset. 

\textbf{Task 2}: Once the DNN $f$ is trained, we propose to add a new task, namely the prediction error estimation. Our goal is to predict the error done by the DNN, i.e., to learn to predict $l_f(f(\mathbf{x}), y)$.
Note that the loss does not need to be the same as $l_f$, since in some cases, $l_f$ could follow a specific design such as in focal loss~\cite{lin2017focal}, in scale-invariant error~\cite{eigen2014depth}, etc. Hence we aim to predict the error loss $l_u(f(\mathbf{x}), y)$. As is the case with regression tasks, a sensible choice for $l_u$ is the mean square or absolute error. Our uncertainty DNN $g$, will have the following training objective: 

\begin{equation}
\setlength\abovedisplayskip{1pt}%shrink space
\setlength\belowdisplayskip{1pt}
\argmin_{g} L\Big( l_u(f(\mathbf{x}), y),g(f(\mathbf{x}),\mathbf{x}) \Big) \label{eq:epsilon}
\end{equation}
where $L$ denotes an objective function which needs to be minimized to obtain a model able to predict the  error of $f$.

Note that the error prediction task of \textbf{task 2}, is related to predicting the total uncertainty of the DNN. The total uncertainty $U(f, \mathbf{x})$ may then be interpreted as the sum of prediction errors according to~\cite{jain2021deup}:
\begin{equation}
\setlength\abovedisplayskip{1pt}%shrink space
\setlength\belowdisplayskip{1pt}
   % U(f,\mathbf{x}) = E[l_u(f(\mathbf{x}), Y)] = \int{l_u(f(\mathbf{x}), y)dP(y|\mathbf{x})}\label{eq:U}
    U(f,\mathbf{x}) = \int{l_u(f(\mathbf{x}), y)dP(y|\mathbf{x})}\label{eq:U}
\end{equation}
In subsection \ref{sec:Architecture}, we explain the structure of the side learner $g$, and in subsection \ref{sec:Loss}, we present the design of the loss $L$ to learn to predict the uncertainty.

\begin{figure}
    \centering{\includegraphics[width=1.0\linewidth]{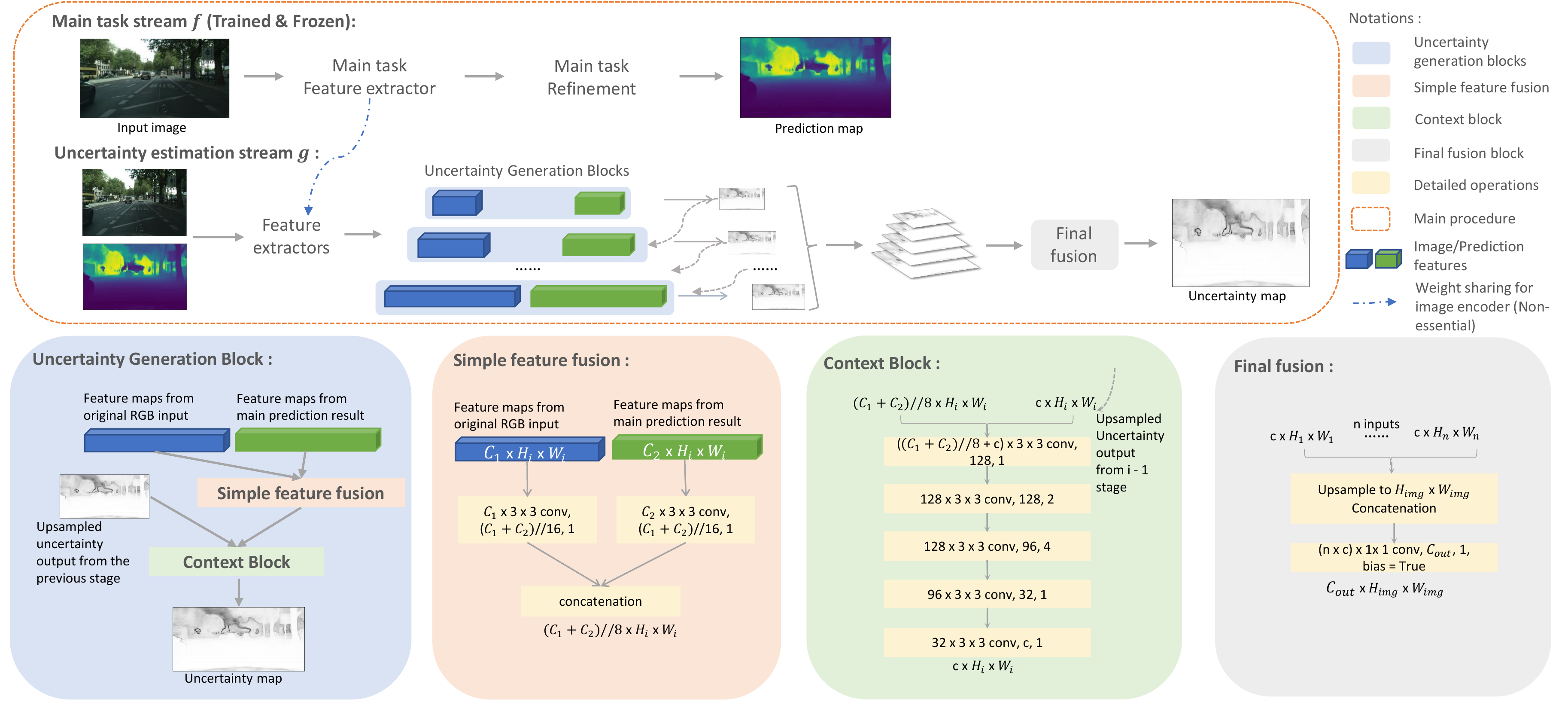}}
    \caption{\small{General solution for pixel-wise uncertainty estimation. We take monocular depth estimation as an example. In main procedure, {\textbf{Main task stream $f$ is first trained and frozen.} Then we use the input and the output of $f$ to train $g$.
    An uncertainty generation block is implemented on feature pairs from different stages. The $n$ outputs of context blocks from different stages will be sent to the final fusion block. The conv-layers in detailed operations are described as (shape, number, dilation ratio).}}}
    \label{fig:general_structure}
\end{figure}

\subsection{Side learner's architecture}\label{sec:Architecture}
The side learner  $g$ takes two inputs:  $\mathbf{x}$ and $f(\mathbf{x})$. The combination of image features and prediction result in uncertainty map that depends on the initial image and on the prediction of $f$.
{The design of this architecture is inspired by empirical observations on the prediction error maps. We notice that, for pixel-wise regression tasks, prediction errors are organized 1. along edges of connected domains in the prediction map; 2. as hard predictable areas which are not captured in the prediction map but only exist in the RGB image, e.g. distant or small objects and occlusions. 
Therefore, the RGB image needs to be combined with the prediction map to recover the semantic information absent from the main task prediction. Meanwhile, we use convolutional features given by the encoders with a final fusion block to better capture the edge information~\cite{liu2017richer}. The concatenated convolutional features are followed by a context block~\cite{sun2018pwc} described as follows.\\}We structure the side learner $g$ in the following three parts, with Figure~\ref{fig:general_structure} showing the general design for our pixel-wise uncertainty estimator.\\ 
\textbf{1. Feature encoders}: They aim to learn and extract the {richer convolutional feature pyramids~\cite{liu2017richer}} from raw input (RGB data) and from the final prediction map preparing for the next steps. We choose the widely used backbone DenseNet~\cite{huang2017densely}. Note that the architecture is agnostic, and DenseNet could be replaced by the other backbones like ResNet~\cite{he2016deep}, depending on the context. {The image encoder in $g$ can be trained \Emi{from scratch} but can also be replaced by the trained and frozen image encoder from $f$. It depends on how closely the image input is related to the prediction error. If we have multiple images as a concatenated input, we need to encode the image on which the prediction error is based. We recommend using the same structure for the feature extractors of the image and the prediction map, so as to \Emi{avoid information asymmetry}.} The following operations will be done from coarse to fine.\\
\textbf{2. Feature fusion}: For each pair of RGB and final prediction features, we concatenate them followed by the convolutional layers in order to reduce the channel number. Dropout layers are implemented before the features are input to the convolutional layer, to reduce the bias caused by focusing on specific patterns. The goal is to find the similarity and the difference between two sources of features to bring out the lost information during main task training.\\
\textbf{3. Context block}: The architecture of this block follows that of the context network in PWC-net~\cite{sun2018pwc}. With various dilation ratios, the convolutional layers can take the features from different size of receptive fields. The output intermediate uncertainty from the previous stage will be up-sampled and concatenated as a {guide} feature to the input of current stage.\\
After obtaining the output of the context block with different resolutions, we up-sample these outputs toward the same size as the prediction target and then use a simple convolutional layer to sum them up with different weights as a final fusion output.
\subsection{Loss design}\label{sec:Loss}
\textbf{1. Natural loss}: A straightforward way is to use mean square error loss (MSE) for $l_u$ in eq.~\ref{eq:epsilon} and eq.~\ref{eq:U}. In order to prevent the distribution of training target from being too sparse, we take $l_u(f(\mathbf{x}),y) = |f(\mathbf{x})-y|$.\\ 
\textbf{2. Target scaling}: The uncertainty of the regression task is expressed across a support based on the current value of the variable, with the support being contained in $\mathbb{R}$. The uncertainty of the classification task is related to the confidence of the classifier output, which is set to take values in the range is $[0, 1]$. Hence we process $l_u(f(\mathbf{x}), y)$ to bound it between 0 and 1.
This can also avoid the influence of the effect from the outliers which have significant prediction errors thanks to the following equation :

\begin{equation} 
\setlength\abovedisplayskip{1pt}%shrink space
\setlength\belowdisplayskip{1pt}
    \tilde{l_u}(f(\mathbf{x}), y) = \tanh(\lambda * l_u(f(\mathbf{x}), y))\label{eq:2}
\end{equation}
where $\lambda$ is a stretch hyper-parameter that can spread the training target as much as possible between 0 and 1, %. Since all transformations are reversible, we can also restore the uncertainty value by using $u^{-1}(f,\mathbf{x})$,
and $\tilde{l_u}(f(\mathbf{x}), y)$ is the normalized uncertainty target.
\\
% \subsubsection{Cross-Entropy Loss}
\textbf{3. Cross-Entropy Loss}: After having normalized the uncertainty $l_u$, {we select cross entropy loss to describe the probability distance between training target $\tilde{l_u}$ and our predicted uncertainty. Since $\tilde{l_u}\in{[0,1]}$, }we choose a binary cross entropy {(BCE)} loss as $L$ in eq.~\ref{eq:epsilon}:

\begin{equation}
\setlength\abovedisplayskip{1pt}%shrink space
\setlength\belowdisplayskip{1pt}\label{eq:4}
\resizebox{.9\hsize}{!}{$
L(\tilde{l_u},g) =  -\sum_i[\tilde{l_u}(f(\mathbf{x}_i),y_i) \log(\sigma(g(f(\mathbf{x}_i),\mathbf{x}_i))) + (1-\tilde{l_u}(f(\mathbf{x}_i),y_i))\log(1-\sigma(g(f(\mathbf{x}_i),\mathbf{x}_i)))]
$}
\end{equation}
%in which subscript $i$ denote pixel point and 
where $\sigma(.)$ is the sigmoid function. {Note that here we use BCE loss not for doing classification, but for regression, and BCE can support a faster convergence for a Sigmoid output value.} By calculating the first derivation of eq.~\ref{eq:4} with respect to $\sigma(g)$, we can find that the optimum will be $\sigma(g) = \tilde{l_u}$. %It can also provide eq.\ref{eq:epsilon} the best $g(f,\mathbf{x})$ by minimizing the binary cross entropy.

\begin{figure}
    \centering{\includegraphics[width=0.8\linewidth]{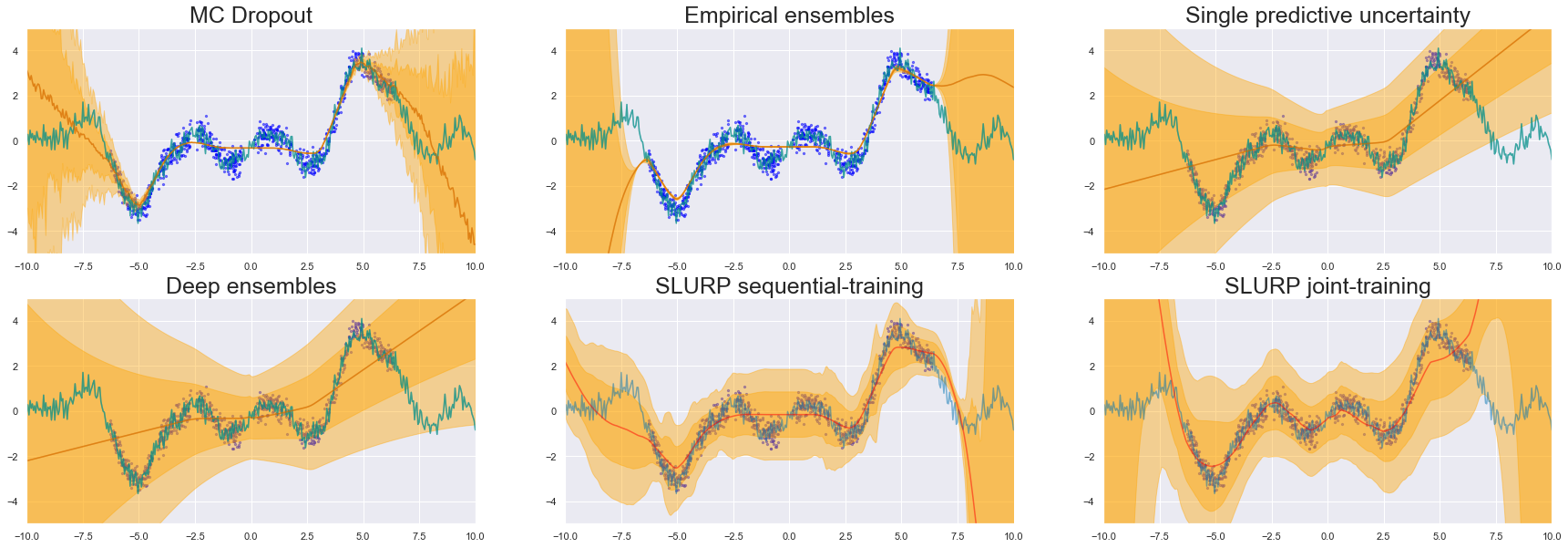}}
    \caption{\small{1D synthetic regression task comparison example. X-axis: spatial coordinate of the Gaussian process. Green curve: ground truth; Blue points: training samples; Red curves: main task prediction; Orange zooms: the uncertainty coverage (1-sigma for inner interval, 2-sigma for outer interval).}}

    \label{fig:toy_dataset}
\end{figure}
\input{expe}
\vspace{-1em}
\section{Conclusion}
\vspace{-0.5em}In this work we proposed a novel solution for estimating the total uncertainty of a DNN. SLURP is a side learner that does not affect the main prediction estimator. Given a trained main task model, we use the original input and main task prediction results as input, and the prediction error as the target. Following feature extraction, fusion and content reconstruction, we train for the total uncertainty of the main task prediction. SLURP can also exploit latent features of the original input provided by a frozen feature extractor of the main task model, such as an image encoder. We compared our proposal with popular uncertainty estimation approaches, and performed detailed experiments on 2D pixel-wise regression tasks to prove that our method is feasible and robust. Future works involve further optimizing the SLURP structure to decrease its footprint and make it accessible for embedded processing, as well as applying this side learning concept to other types of input signals. 
\clearpage
\bibliography{egbib}

%%%%%%%%%% Merge with supplemental materials %%%%%%%%%%
\clearpage
\hspace*{\fill}
\begin{center}
\textbf{\large SLURP: Side Learning Uncertainty for Regression Problems \\--- Supplementary Material ---}
\end{center}
\hspace*{\fill}

\input{bmvc_supp}
\input{expe_supp}
\clearpage
\end{document}

%% file: intro.tex
%------------------------------------------------------------------------- 
\section{Introduction}
The increasing use of Deep Neural Networks (DNNs) fuelled significant advances in the last decade in Computer Vision and fully benefited from the improvements in data availability and computational power. Currently, the main factor impeding a wider adoption of DNNs for critical tasks is their lack of reliability and interpretability. Subsequently, a lot of effort has been dedicated to the uncertainty estimation of model predictions during inference, which aims to alleviate the impact of out-of-distribution and noisy inputs. The most popular approaches are either based on ensembles, or on some simplificatory assumptions applied to Bayesian Neural Networks~\cite{gal2016dropout, lakshminarayanan2016simple,batch}. In both cases, the predictive uncertainty is related to the output diversity, the quality of which is correlated to a higher computational cost. Moreover, classification models have received more attention since their output is simpler and allows for an easier definition of uncertainty estimation metrics~\cite{hendrycks2016baseline,hendrycks2019benchmarking}. In a regression setting, DNNs are expected uncertainty-wise to provide a probability distribution over the target domain, but most existing models also lack a default uncertainty output. Following the works addressing classification models, some strategies and metrics for uncertainty estimation and calibration have been proposed more recently for regression~\cite{ilg2018uncertainty,gustafsson2020evaluating}. However, the proposed solutions tend to be more specific due to the higher variability of the output domain associated with diverse input data, \emph{e.g} time series forecasting vs. pixel-wise regression. %The proposed solutions however tend to be more specific due to the higher variability of the output domain, associated to diverse input data, e.g. time series forecasting vs. pixel-wise regression.
As a default solution, ensemble-based strategies are relatively effective but incur a significant training and inference computational cost, which is sometimes prohibitive when considered along with the memory requirements for some tasks.

Alternatively, using a side learner~\cite{jain2021deup,yoo2019learning,hu2020new, corbiere2019addressing} may provide a generic solution for predictive uncertainty estimation.
The side learner could be regarded as a post-processing applied to the prediction result. Therefore, it is a straightforward approach to provide a precise uncertainty estimation without influencing the performance of the main task model or further refining its hyper-parameters. However, for some pixel-wise regression tasks, this approach may be challenging because of the importance of the semantic context.
%The side learner could be regarded as a post-processing applied to the prediction result, and it is therefore a straightforward approach to provide a precise uncertainty estimation without influencing the performance of / requiring efforts for  modifying the main task model or further refining its hyper-parameters. For some pixel-wise regression tasks however, this approach may be challenging because of the importance of the semantic context. 
Compared with the uncertainty for time series data that pays more attention to long short-term information, the uncertainty of image-based tasks should give more consideration to the semantics of the image. The uncertainty map is based on the prediction map because of the similar higher level encoded semantics. However, obviously, the prediction result might miss relevant semantics with respect to the ground truth, which may have a detrimental effect on the  uncertainty estimation. Our work intends to fill this gap by introducing SLURP, a general side learning approach for regression problems able to recover semantic information absent from the main task prediction. 
Our contributions are as follows.
{\textbf{1. }We are the first to solve the uncertainty estimation problem for general pixel-wise regression tasks with an auxiliary network. \textbf{2. }We propose a transposable architecture which may be used along with the main task model without modifying/re-training/affecting the performance of the latter, thus greatly improving our proposal's adoption potential. \textbf{3. }We demonstrate our side learner's flexibility on two fundamental vision tasks.  The extensive experiments validate our algorithm's consistent performance in line with SOTA uncertainty estimation algorithms. \Xuanlong{The main advantage of our proposed solution is the efficiency and simplicity of its implementation and its competitive uncertainty performance.}}

%% file: related.tex
\section{Related Works}
\begin{figure}
    \centering{\includegraphics[width=1.0\linewidth]{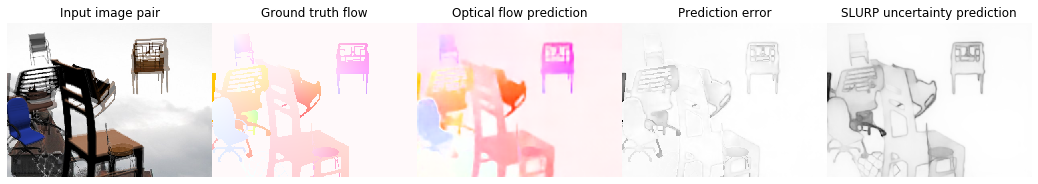}}
    \caption{\small{An example for uncertainty estimation on optical flow task. The predicted flow is made by FlowNetS~\cite{dosovitskiy2015flownet}. The prediction error is the end-point-error between ground truth flow and predicted flow. We can see that our uncertainty estimation map can correspond well to the true error, {including the semantic loss area (middle left) and the edges in the connected domains of the prediction flow map.}}}
    \label{fig:vis_optical_flow}
\end{figure}
\subsection{Major uncertainty estimation approaches}
\textbf{1. Distribution-based approaches} By assuming a probability distribution over the model output and by minimizing the associated negative log-likelihood{~\cite{Nix1994EstimatingTM, kendall}}, one may obtain the aleatoric uncertainty of the prediction result. Based on this idea,~\cite{asai2019multi} separate the formulation and use it as a two-task learning which could improve the main task and uncertainty performance on indoor depth estimation. As for the considered distribution, Gundavarapu~\etal~\cite{gundavarapu2019structured} choose to use a multi-variate Gaussian to capture the aleatoric uncertainty in human pose, while in their experiments~\cite{gast2018lightweight,ilg2018uncertainty} choose the best performing distribution from an exponential distribution cluster. Apart from supervised methods,~\cite{poggi2020uncertainty} learns uncertainty from monocular depth in a self-supervised way with a teacher-student framework using the concept of~\cite{kendall}, while being too specific for re-implementation on the other tasks. \\
\textbf{2. Ensemble-based approaches} In the statistics community~\cite{efron1994introduction,steck2003bias} and later in computer vision~\cite{kybic2011bootstrap,ilg2018uncertainty}, the most conventional way to capture model uncertainty has been to use bootstrapping in which the uncertainty is linked to the variance of multiple outputs generated by re-sampling. Although it is accessible and well studied, this approach is not well suited to deep learning as it tends to degrade the main task performance. Instead of using part of the data each time, Deep Ensembles~\cite{lakshminarayanan2016simple} uses the full data to train the model under different stochasticity sources and combines the aleatoric uncertainty estimation proposed by~\cite{kendall}. %It then provides a mixture uncertainty. 
While this method is easily adaptable to different tasks while still achieving an extremely competitive baseline~\cite{ashukha2020pitfalls}, its memory and run-time consumption make it inefficient in real applications due to its dependence on multiple, potentially heavy models. Eddy Ilg \etal~\cite{ilg2018uncertainty} use a multiple-hypothesis network combining with the concept of~\cite{kendall} to reduce memory consumption of ensembles. However, {it is specific for optical flow, it has to modify the main task model and add an extra network in the end for merging the hypotheses, and it requires multiple stages of training.} \\
\textbf{3. Bayesian approaches} Learning the posterior distribution of the weights of the main task neural network given the training dataset is a way to obtain the uncertainty from the network architecture~\cite{blundell2015weight}. Gast \etal~\cite{gast2018lightweight} try to weight the uncertainty from network weight by using assumed density filtering~\cite{boyen2013tractable}, but in this case the network has to be modified. MC-Dropout~\cite{gal2016dropout} is a widely used method which mimics a BNN by using injected Dropout layers to sample the network weights. It needs to do multiple times forward propagation which is costly especially for pixel-wise regression tasks. At the same time, an additional challenge is raised by properly balancing the main performance and the quality of the generated uncertainty due to the number and position of the injected Dropout layers.

\subsection{Side learning for uncertainty estimation}
With the development of deep learning, DNNs are becoming more sophisticated, and the training of DNN and the selection of hyperparameters have also become particularly important~\cite{sun2019models}. As it avoids changing the main task model structure, a side learner is in this respect a convenient approach to estimate uncertainty. {Lee~\etal~\cite{lee2020fast} use a conditional GAN~\cite{isola2017image} as an auxiliary model to project an image-pair input in optical flow to an uncertainty provided by a MC-Dropout version of main task to reduce the time cost, but its performance will be decreased during dataset shift.}
Yoo~\etal~\cite{yoo2019learning} design a side learner to learn the loss supplied by the main task model. {However, the goal of this work is to provide \textit{a single predicted loss} for the main task, which is not suitable for pixel-wise uncertainty estimation.}
%Narayanaswamy \etal~\cite{narayanaswamy2021loss} also implement a side learner to predict loss, and they give an insight that the side loss estimator could improve the main task performance but they only work on classification tasks.
Instead of predicting the loss from the main task model, based on~\cite{lakshminarayanan2016simple} and~\cite{kendall}, Hu~\etal~\cite{hu2020new} 
% wants to fit a side learner to predict the prediction error of the optimal mean prediction obtained by an ensemble.
{trains sequentially a network which is \Emi{identical} to the main task \Emi{model} to fit the prediction error.
%generated by an ensemble of main task networks to
It can \Emi{improve} the stability of uncertainty training but the memory cost of the uncertainty estimator will increase with the increase of the main task model scale.} DEUP~\cite{jain2021deup} not only uses a side learner to predict the loss but in order to better obtain the epistemic uncertainty, it enriches the input of the side learner
%, using the original data point, the predicted aleatoric uncertainty, the identifier of whether it has been trained and data point density. I, and it shows a very good performance on various tasks, 
but it needs to train three models including a main task model with aleatoric uncertainty estimator, a data density estimator and a loss side learner. {This has high requirements for model selection and training.} {Corbière~\etal~\cite{corbiere2019addressing} introduces ConfidNet which can conveniently provide the uncertainty by adding an auxiliary network after the main task encoder. It shares a more similar approach with our work. It shows a good performance but it works only for classification task and it has to re-tune the encoder part for uncertainty prediction so it structurally depends on the main task network and requires two-stage training.}\\
For pixel-wise regression tasks, the number of data points is huge and data preparation will be costly. SLURP does not need to modify the main task model, does not require data preparation. {It faces on general pixel-wise regression tasks, which is different from all the works mentioned above. We use a direct and explicit design, and only train the side learner once to get the uncertainty. This can circumvent the difficulty of modifying the main task model caused by the complexity of the structure, or the lack of the training codes.} Using only the in-distribution data, without touching the main task model, we can get better quality and robust uncertainty. {Fig.~\ref{fig:vis_optical_flow} gives an example on uncertainty estimation of our work.}

%% file: expe.tex
\section{Experiments}
\vspace{-0.5em}Our main focus is on obtaining high-quality uncertainty maps on pixel-level regression tasks. However, in order to illustrate more comprehensively our approach and show its applicability in a different context, we also apply SLURP on a 1D toy dataset. Overall, our proposed method is illustrated on a synthetic 1D regression dataset, and on two fundamental computer vision tasks: optical flow (OF) and monocular depth (MD). For the former, we just visualize it to give some reference and insights. For the later ones, we evaluate the quality of predicted uncertainty maps, the basic idea is to see whether the predicted uncertainty map matches the prediction error. To this end, we re-implemented the transferable uncertainty estimation approaches MC-Dropout (MC)~\cite{gal2016dropout}, Single predictive uncertainty (Single-PU)~\cite{kendall}, Empirical ensembles (EE){, ConfidNet (Confid)~\cite{corbiere2019addressing} and} Deep ensembles (DE)~\cite{lakshminarayanan2016simple} to the main task as the comparisons. {Due to its particularity, we reproduce the multi-hypothesis prediction network (MHP)~\cite{ilg2018uncertainty} only for the optical flow task. Specifically, to transfer Confid solution from the classification task to the regression task, we duplicate the last few layers from the last de-convolutional layer (or up-sample operation) and add three extra 3x3 convolutional layers without changing the resolution as ConfidNet for regression. We keep the training schedule, and the pixel-wise square error maps will replace the original training targets.} We use two evaluation criteria: AUSE and AUROC to evaluate uncertainty maps generated by different methods. {Additionally, we measure the efficiency of main task - uncertainty estimator system from two perspectives: Runtime and Number of parameters.} The specific formulas as well as the implementation details for the methods we compare with are provided in the supp. material.

\vspace{-1em}\subsection{1D regression task toy example}
We compare our proposed solution with MC~\cite{gal2016dropout}, EE, Single-PU~\cite{kendall} and DE~\cite{lakshminarayanan2016simple} on a 1D regression task dataset. The toy dataset is generated by Gaussian process, our spatial coordinate range is $x_i \in{[-10, 10]}$. From $x_i \in{[-7, 7]}$, we cross-select 875 data points as training set and 175 data points as validation set. From $x_i \in{[-10, 10]}$, we randomly select 400 points as test set. The main task predictor $f$ with single output has only one hidden layer, the Single-PU and DE are implemented based on a modified dual-output $f$. The detailed model and training settings for all the methods are in the supp. material. Moreover, we use two modes to train our SLURP side learner $g$. One is that we first train one $f$, then we freeze it and train the $g$ with the prediction result from $f$ and the latent feature extracted from the single hidden layer in $f$, i.e., sequential-training{, which is the original design of our approach}. Another one is that we train $f$ and $g$ at the same time by using the negative log-likelihood loss~\cite{kendall}, i.e., joint-training. Because of the data dimension, the side learner shown in Fig~\ref{fig:general_structure} is modified (see supp. material). As illustrated in Fig~\ref{fig:toy_dataset}, the results of MC~\cite{gal2016dropout} and EE give a good uncertainty on the unseen area, but little on the training part. While DE~\cite{lakshminarayanan2016simple} and Single-PU~\cite{kendall} can give a sufficient uncertainty to all areas. But in comparison, their main task prediction accuracy has been affected. SLURP can achieve both reasonable main task accuracy and tight uncertainty coverage especially for joint-training one.
\vspace{-1em}\subsection{Evaluation protocols for pixel-wise tasks}\label{eval_matrics}
\vspace{-0.2em}MD and OF are two fundamental regression tasks which have significant implications in a wide range of applications. In terms of inputs, the main difference is that MD requires a single RGB image, while OF needs an image pair. For each pixel, the MD output is a depth value $y_d \in \mathcal{R}^+$, while the OF output is a 2-channel displacement vector $y_o \in \mathcal{R}^2$. We introduce below the evaluation criteria.\\
\textbf{1. Uncertainty ordering}: Let us consider that we want to remove the worst pixels based on an uncertainty estimator; we expect that a good uncertainty estimator should allow us to remove the less reliable data. This is evaluated thanks to the \textbf{sparsification curve (SC)}, and the \textbf{area under sparsification error (AUSE})~\cite{bruhn2006confidence, kondermann2008statistical, kybic2011bootstrap, ilg2018uncertainty, poggi2020uncertainty}.
To build the SC, given a set of data and their uncertainty, we iteratively erase $m\%$ (we take $m=5$ in our experiments) of the data which exhibit the highest uncertainty. Then we calculate the average prediction error for the remaining data. Hence we have the SC. To  evaluate the oracle SC, we remove the $m\%$ data with the most significant prediction error  and we calculate the average prediction error for the remaining data.
We denote the area between the two SC curves as AUSE. The smaller the AUSE, the closer the order of the predicted uncertainty and the order of the Oracle. As a note, AUSE could be changed if we change the %loss criterion
error metric. Therefore, we denote AUSE based on different %$\tau$
error metrics as AUSE-xxx.
\\\textbf{Implementations}: Specifically, for monocular depth, we choose square error and absolute relative error~\cite{eigen2014depth} .The corresponding AUSEs are \textbf{AUSE-RMSE} and \textbf{AUSE-Absrel}. For optical flow we use EPE, which is the error map representing the Euclidean distance between the ground truth motion and the predicted one, we denote its AUSE as \textbf{AUSE-EPE}. \\
\textbf{2. AUROC}: Since we have access to a soft evaluation of uncertainty, it is feasible to threshold the dataset into two sets, namely the reliable set and the unreliable set. We propose for MD to set the data as reliable if they check the inlier metrics threshold d1 criterion proposed in~\cite{hu2017robust},
and for OF data is reliable if its EPE is below $k=2$.
We scale the predicted uncertainty between 0 and 1 with min-max scaling and evaluate the ROC curve.
The larger the AUROC, the more data points are given correct confidence (uncertainty) and the better the uncertainty estimator is.\\
{\textbf{3. Model efficiency:} Due to the different schemes of uncertainty generation design, we will measure the number of model parameters and time consumption of the entire system (\textbf{\# Param.}), including the main task model and uncertainty generator. We count the running time (\textbf{Runtime}) while processing a whole testing dataset in one NVIDIA TITAN RTX GPU and Intel Core i9-10900X CPU then take the average according to the number of samples.}
\vspace{-1em}\subsection{Monocular depth}
In this section, we introduce 
% the implementation details on
uncertainty estimations for MD. We choose BTS~\cite{lee2019big} as the depth estimator, and the uncertainty estimators will be implemented based on this architecture. BTS is one of the state-of-the-art architectures on MD benchmarks~\cite{silberman2012indoor, geiger2013vision}. As an encoder-decoder based network it is well suited for the extraction of latent image features as shown in Fig~\ref{fig:general_structure}, however note that our strategy is agnostic to the main task architecture. %We use it as an example for uncertainty estimation of the latest architectures. 
In accordance with the default setting, we choose DenseNet161~\cite{huang2017densely} as encoder {for BTS}. {So for our side learner, we take the trained and fixed image encoder of BTS as our image encoder, and implement a new DenseNet161 encoder for estimated depth map.}
\\
\noindent\textbf{Datasets and procedures}
We choose two widely used datasets and their variations to train and evaluate uncertainty estimators as follows. \textit{Training set:} KITTI~{\cite{geiger2013vision, Uhrig2017THREEDV}} Eigen-split training set~\cite{eigen2014depth}; \textit{Fine-tuning set:} Cityscapes training set~\cite{cordts2016cityscapes}; \textit{Test set:} KITTI Eigen-split test set, Cityscapes test set, Foggy Cityscapes-DBF test set~\cite{SDHV18} with three severity levels and Rainy Cityscapes test set~\cite{hu2019depth} with three parameter sets which indicate three severities. KITTI depth dataset and Cityscapes dataset are both outdoor datasets. However, since the ground truth on KITTI is sparse and has only half the content of the image, the model performance is limited if training on KITTI and testing on Cityscapes. Therefore, we first train models on KITTI Eigen-split training set then evaluate the uncertainty on KITTI Eigen-split test set and Cityscapes test set. After that, we fine-tune the models on Cityscapes training set and evaluate the uncertainty on all Cityscapes test sets listed above.
\noindent\textbf{Results}
Table~\ref{table:result_depth} presents the performance of different methods in normal circumstances and against gradual input perturbations. The top performing method is highlighted in bold, while the second one is highlighted in blue. SLURP, ConfidNet and Deep ensembles exhibit competitive performance being both ranked close in terms of uncertainty ordering on the different metrics, with our proposal being clearly more robust against strong input perturbations. 

\begin{table}[t!]
\small
\begin{center}\resizebox{0.6\columnwidth}{!}{%
\begin{tabular}{ccccccc||c} 
\hline
Datasets & Criteria & MC & EE & Single PU & DE & Confid & Ours \\ 
\hline
\multirow{3}{*}{KITTI} & AUSE-RMSE & 8.14 & 3.17 & \textcolor{blue}{1.89} & \textbf{1.68} & 1.76 & \textbf{1.68} \\
 & AUSE-Absrel & 9.48 & 5.02 & 4.59 & \textcolor{blue}{4.32} & \textbf{4.24} & 4.36 \\
 & AUROC & 0.686 & 0.882 & 0.882 & \textbf{0.897} & 0.892 & \textcolor{blue}{0.895} \\ 
\hline
\multirow{3}{*}{CityScapes} & AUSE-RMSE & \textbf{9.42} & 11.56 & 9.91 & 11.47 & 10.48 & \textcolor{blue}{9.48} \\
 & AUSE-Absrel & 9.52 & 13.14 & 9.96 & \textcolor{blue}{9.36} & \textbf{5.75} & 10.90 \\
 & AUROC & 0.420 & \textcolor{blue}{0.504} & 0.386 & 0.501 & \textbf{0.519} & 0.400 \\ 
\hline
\multicolumn{8}{c}{After fine-tuning on CityScapes} \\ 
\hline
\multirow{3}{*}{CityScapes} & AUSE-RMSE & 7.72 & 8.20 & 4.35 & \textbf{3.03} & 4.05 & \textcolor{blue}{3.05} \\
 & AUSE-Absrel & 8.13 & 7.50 & \textcolor{blue}{6.44} & 6.81 & \textbf{6.34} & 6.55 \\
 & AUROC & 0.705 & 0.786 & 0.741 & \textbf{0.856} & 0.821 & \textcolor{blue}{0.849} \\ 
\hline
\multirow{3}{*}{\begin{tabular}[c]{@{}c@{}}CityScapes\\Rainy s=1\end{tabular}} & AUSE-RMSE & 7.06 & 7.29 & 4.17 & \textcolor{blue}{3.42} & 4.89 & \textbf{3.39} \\
 & AUSE-Absrel & 8.73 & 6.92 & \textcolor{blue}{6.55} & 6.68 & 7.26 & \textbf{5.62} \\
 & AUROC & 0.659 & \textcolor{blue}{0.757} & 0.731 & 0.746 & 0.697 & \textbf{0.788} \\ 
\hline
\multirow{3}{*}{\begin{tabular}[c]{@{}c@{}}CityScapes\\Rainy s=2\end{tabular}} & AUSE-RMSE & 7.14 & 6.9 & 4.27 & \textbf{3.35} & 4.68 & \textcolor{blue}{3.36} \\
 & AUSE-Absrel & 8.36 & 6.48 & 6.79 & \textcolor{blue}{6.24} & 6.86 & \textbf{5.28} \\
 & AUROC & 0.667 & \textcolor{blue}{0.767} & 0.731 & 0.756 & 0.714 & \textbf{0.794} \\ 
\hline
\multirow{3}{*}{\begin{tabular}[c]{@{}c@{}}CityScapes\\Rainy s=3\end{tabular}} & AUSE-RMSE & 7.30 & 6.66 & 4.35 & \textbf{\textbf{3.28}} & 4.59 & \textcolor{blue}{3.41} \\
 & AUSE-Absrel & 8.27 & 6.03 & 6.44 & \textcolor{blue}{5.85} & 6.64 & \textbf{5.05} \\
 & AUROC & 0.665 & \textcolor{blue}{0.778} & 0.742 & 0.767 & 0.729 & \textbf{0.801} \\ 
\hline
\multirow{3}{*}{\begin{tabular}[c]{@{}c@{}}CityScapes\\Foggy s=1\end{tabular}} & AUSE-RMSE & 7.80 & 7.82 & 3.42 & \textcolor{blue}{3.05} & 3.98 & \textbf{3.04 } \\
 & AUSE-Absrel & 8.36 & 7.33 & 6.78 & 6.58 & \textbf{6.21} & \textcolor{blue}{6.25} \\
 & AUROC & 0.700 & 0.783 & 0.842 & \textbf{0.852} & 0.824 & \textcolor{blue}{0.847} \\ 
\hline
\multirow{3}{*}{\begin{tabular}[c]{@{}c@{}}CityScapes\\Foggy s=2\end{tabular}} & AUSE-RMSE & 7.82 & 7.53 & 3.42 & \textbf{2.98} & 3.86 & \textcolor{blue}{3.01} \\
 & AUSE-Absrel & 8.20 & 7.09 & 6.55 & 6.35 & \textbf{6.02} & \textcolor{blue}{6.06} \\
 & AUROC & 0.704 & 0.791 & 0.847 & \textbf{0.857} & 0.833 & \textcolor{blue}{0.852} \\ 
\hline
\multirow{3}{*}{\begin{tabular}[c]{@{}c@{}}CityScapes\\Foggy s=3\end{tabular}} & AUSE-RMSE & 7.84 & 7.28 & 3.48 & \textbf{2.93} & 3.70 & \textcolor{blue}{3.08} \\
 & AUSE-Absrel & 7.87 & 6.80 & 6.19 & 6.01 & \textbf{5.78} & \textcolor{blue}{5.80} \\
 & AUROC & 0.715 & 0.801 & 0.851 & \textbf{0.863} & 0.846 & \textcolor{blue}{0.857} \\
\hline
\end{tabular}
}
\end{center}
\caption{\small{MD uncertainty performance. Bold value: result with the best performance, Blue value: second performance. s (\emph{e.g} s=1) indicates severity, higher the s value, higher the severity.}}
\label{table:result_depth}
\end{table}

\vspace{-1em}\subsection{Optical flow}
FlowNetS~\cite{dosovitskiy2015flownet} is a  popular OF architecture, as the first learning-based optical flow estimation method. We apply it as our main task model and implement the uncertainty estimators based on it, since it can be regarded as a good example for uncertainty estimation on early structures with special inputs. FlowNetS is also an encoder-decoder network, however - differently from the more recent architectures with a separate feature extractor implemented for the first image~\cite{ilg2017flownet, sun2018pwc,hui2018liteflownet,yang2019volumetric, poggi2020uncertainty}, FlowNetS takes directly image pairs as encoder input, so the features from different encoder stages will be mixture feature maps of two RGB images. Since the ground truth motion is based on the first image, the true error will be based only on the first image, we use a new encoder to extract the RGB features only for the first image. {In this case, in our side learner, we use two trainable encoders respectively for the first image and the predicted flow. We choose both DenseNet161 and DenseNet121~\cite{huang2017densely} as the encoders in our experiments and we denote the latter Ours-light as a lightweight version of the former.}
\noindent\textbf{Datasets}
\textit{Training set:} synthetic FlyingChairs training set~\cite{dosovitskiy2015flownet}; \textit{Testing set:} FlyingChairs test set, synthetic Sintel training set~\cite{Butler:ECCV:2012} and real-world KITTI {2015 training set~\cite{geiger2013vision, Menze2015ISA, Menze2018JPRS}}. The train-test split file for FlyingChairs dataset is provided officially. Sintel has more complex moving objects and movements than FlyingChairs, while KITTI is taken from real-world, and exhibits larger movement magnitude.
\begin{table}[t!]
\small
\begin{center}
\resizebox{0.7\columnwidth}{!}{%
\begin{tabular}{cccccccc||cc} 
\hline
Datasets & Criteria & MC & EE & Single PU & DE & Confid & MHP & Ours-light & Ours \\ 
\hline
\multirow{2}{*}{FlyingChairs} & AUSE-EPE & 2.75 & 1.97 & 1.28 & 1.26 & 1.92 & 1.88 & \textbf{1.16} & \textcolor{blue}{1.20} \\
 & AUROC & 0.896 & 0.900 & \textbf{0.977 } & 0.959 & 0.945 & 0.936 & \textbf{0.977} & \textcolor{blue}{0.974} \\ 
\hline
\multirow{2}{*}{KITTI} & AUSE-EPE & \textcolor{blue}{3.57 } & 4.41 & 3.71 & \textbf{3.45 } & 6.56 & 5.48 & 5.20 & 4.69 \\
 & AUROC & \textcolor{blue}{0.870 } & \textbf{0.904} & 0.848 & 0.866 & 0.687 & 0.854 & 0.791 & 0.800 \\ 
\hline
\multirow{2}{*}{Sintel Clean} & AUSE-EPE & 3.33 & 2.89 & \textcolor{blue}{2.74 } & 3.02 & 5.28 & \textbf{2.61} & 3.02 & 2.91 \\
 & AUROC & 0.861 & 0.825 & \textbf{0.925 } & 0.895 & 0.767 & 0.886 & 0.889 & \textcolor{blue}{0.896} \\ 
\hline
\multirow{2}{*}{Sintel Final} & AUSE-EPE & 3.30 & 3.02 & 3.09 & 3.05 & 6.06 & \textbf{2.71} & 2.95 & \textcolor{blue}{2.86} \\
 & AUROC & 0.858 & 0.814 & \textbf{0.916 } & 0.899 & 0.728 & 0.878 & 0.901 & \textcolor{blue}{0.906} \\
\hline
\end{tabular}
}
\end{center}
\caption{\small{OF uncertainty performance. Bold value: result with the best performance. Blue value: second performance.}}
\label{table:result_optical}
\end{table}

\noindent\textbf{Results}
Table~\ref{table:result_optical} shows the uncertainty estimation performance on the various OF datasets. Even though all uncertainty estimators are not fine-tuned on datasets other than the training set, they maintain a good estimation level. On this very different scenario, MHP did a good job due to its pertinence, but our side learner performs competitively and robustly across all tests/metrics. 
{\vspace{-1em}\subsection{Model efficiency}
\textbf{Datasets and settings }We select to use Sintel training set~\cite{Butler:ECCV:2012} and KITTI~\cite{geiger2013vision} Eigen-split validation set~\cite{eigen2014depth} as the testing sets for Runtime evaluation. We use three models to form EE and DE and eight forward propagations for MC. In MD, due to the instability of training, we use sequential-training for Single-PU and DE. More details are in the supp. material.\\
\textbf{Results } Table~\ref{table:consumption} shows the model efficiency in two tasks. In OF task, due to the convenience of modifying the main task model, we confirm that Single-PU is more efficient than the other methods, while our lightweight version can achieve comparable performance. For MD, due to the complexity in implementation for joint uncertainty generation, the side-learner-based methods have a fixed computational footprint and take significant advantage in the cost. In addition, our solution can reuse without re-tuning the main task encoder, so it is lighter and faster than the other auxiliary networks.  
\begin{table}[!t]
\begin{center}
\resizebox{0.7\columnwidth}{!}{%
\begin{tabular}{cccccccc||cc} 
\hline
Task & Criteria & MC & EE & Single-PU & DE & Confid & MHP & Ours-light & Ours \\ 
\hline
\multirow{2}{*}{MD} & Runtime (ms) & 386 & 144 & \textcolor{blue}{98} & 286 & 106 & - & - & \textbf{88} \\
 & \# Param. (M) & \textbf{47.0} & 141.0 & 94.0 & 282.0 & 94.7 & - & - & \textcolor{blue}{87.2} \\ 
\hline
\multirow{2}{*}{OF} & Runtime (ms) & 79 & \textcolor{blue}{65} & \textbf{\textbf{64}} & 66 & \textcolor{blue}{65} & 67 & \textcolor{blue}{65} & 76 \\
 & \# Param. (M) & \textbf{\textbf{38.7}} & 116.1 & \textbf{\textbf{38.7}} & 116.3 & 78.6 & 78.8 & \textcolor{blue}{57.0} & 105.3 \\
\hline
\end{tabular}
}
\end{center}
\caption{\small{Average time cost for processing one image and number of parameters of the model(s) (main task + uncertainty task). Bold value: result with the best performance. Blue value: second performance.}}
\label{table:consumption}
\end{table}
}

%% file: bmvc_supp.tex
\newcommand{\xmark}{\ding{55}}%
\newcommand{\cmark}{\ding{51}}%
\setlength{\textfloatsep}{8pt}
% \title{SLURP: Side Learning Uncertainty for Regression Problems \\ \hfill--- Supplementary Material ---\hfill}

% Enter the paper's authors in order
% \addauthor{Xuanlong Yu}{xuanlong.yu@universite-paris-saclay.fr}{1}
% \addauthor{Gianni Franchi}{gianni.franchi@ensta-paris.fr}{2}
% \addauthor{Emanuel Aldea}{emanuel.aldea@universite-paris-saclay.fr}{1}

% Enter the institutions
% \addinstitution{Name\\Address}
% \addinstitution{
%  SATIE, Paris-Saclay University\\
%  Gif-sur-Yvette, France
% }
% \addinstitution{
%  U2IS, ENSTA Paris\\
%  Institut Polytechnique de Paris\\
%  Palaiseau, France
% }

\runninghead{Yu et al.}{Side Learning Uncertainty for Regression Problems with SLURP}
% Any macro definitions you would like to include
% These are not defined in the style file, because they don't begin
% with \bmva, so they might conflict with the user's own macros.
% The \bmvaOneDot macro adds a full stop unless there is one in the
% text already.
\def\eg{\emph{e.g}\bmvaOneDot}
\def\Eg{\emph{E.g}\bmvaOneDot}
\def\etal{\emph{et al}\bmvaOneDot}

%------------------------------------------------------------------------- 
% Document starts here
% \begin{document}

The document is structured as follows. We recall first the abbreviations which are used and which will help the reader follow the next sections (Section \ref{sec:notations}). Then, we present and discuss the ablation study (Section \ref{sec:ablation}). Section \ref{sec:expe} provides all the information which is needed in order to duplicate and evaluate the result of the experiments performed in the main paper, more specifically the monocular depth estimation (Section \ref{ssec:MDE_settings}), the optical flow estimation (Section \ref{ssec:of}) and the toy example (Section \ref{ssec:toy}). Information about the accuracy of the main task models is included as well. Finally, Section \ref{sec:moar} presents some additional qualitative results which provide further insights about the behavior of SLURP and of the other considered strategies.

\section{Notations}
\label{sec:notations}

In Table~\ref{tab:abbraviations2} we summarize the abbreviations used in the paper.
\begin{table}
[h!]
\small
\begin{center}\resizebox{0.8\columnwidth}{!}{
\begin{tabular}{ll} 
\hline
\textbf{Abbreviation} & \textbf{Meaning} \\ 
\hline
AUROC & Area under receiver operating characteristic curve \\ 
\hline
AUSE & Area under sparsification error \\ 
\hline
Confid & ConfidNet uncertainty estimation solution \\ 
\hline
DE & Deep ensembles \\ 
\hline
EE & Empirical ensemble \\ 
\hline
MC & MC-Dropout \\ 
\hline
MD & Monocular depth \\ 
\hline
MHP & Multi-hypothesis prediction network for optical flow uncertainty estimation\\ 
\hline
OF & Optical flow \\ 
\hline
SC & Sparsification curve \\ 
\hline
Single-PU & Single predictive uncertainty \\ 
\hline
 & 
\end{tabular}
}
\caption{\small{Summary of the abbreviations of the paper.}}
\label{tab:abbraviations2}
\end{center}
\end{table}

% \end{document}

%% file: expe_supp.tex
\section{Ablation study}
\label{sec:ablation}
\textbf{1. Ablation study settings}: The ablation study for the monocular depth (MD) task is implemented on KITTI~\cite{geiger2013vision, Uhrig2017THREEDV} Eigen-split test set~\cite{eigen2014depth}, Cityscapes test set~\cite{cordts2016cityscapes}, Foggy Cityscapes-DBF test set~\cite{SDHV18} and Rainy Cityscapes test set~\cite{hu2019depth}. The ablation study for the optical flow (OF) task is implemented on FlyingChairs test set~\cite{dosovitskiy2015flownet}, Sintel training set~\cite{Butler:ECCV:2012} and KITTI 2015 training set~\cite{geiger2013vision, Menze2015ISA, Menze2018JPRS}. Brief descriptions of these datasets can be found in the main paper. We use the same uncertainty evaluation metrics (AUSE and AUROC) as section 4.2 in the main paper.\\\\
\textbf{2. Ablation study goals}: We want to highlight the impact of the two considered inputs on the final performance (namely the image features and the prediction results features), and the impact of the considered loss (binary cross entropy loss and mean square error loss). 
\\\\
\textbf{3. Results}: The models with different inputs and different loss functions are presented as follows. Table~\ref{tab:ablation_of} presents the model performance on OF task and Table~\ref{tab:ablation_MDE} illustrates the results on MD task. Note that in the tables BCE and MSE denote binary cross entropy loss and mean square error loss respectively,  PredOnly and RGBOnly denote the models taking only prediction map as input and the models taking only RGB image as input respectively. No special note means that the model will use both RGB and prediction results as input and BCE as the loss function (the default behavior).\\\\
\textbf{4. Discussions}: Firstly, regarding the performance of the different loss functions, we found that the results obtained with the BCE loss are almost systematically better than those provided when using MSE loss. We think this is because when we have a correctly trained predictor for the main task, most of the data points have minor errors, while a small number of data points have high errors. Using the MSE loss will amplify the more significant prediction errors and reduce the minor errors, making the model unable to fit well. Our target scaling uses a soft clipping strategy to centralize the distribution of data for better fitting.

\begin{figure}[t!]
    \centering{\includegraphics[width=1\linewidth]{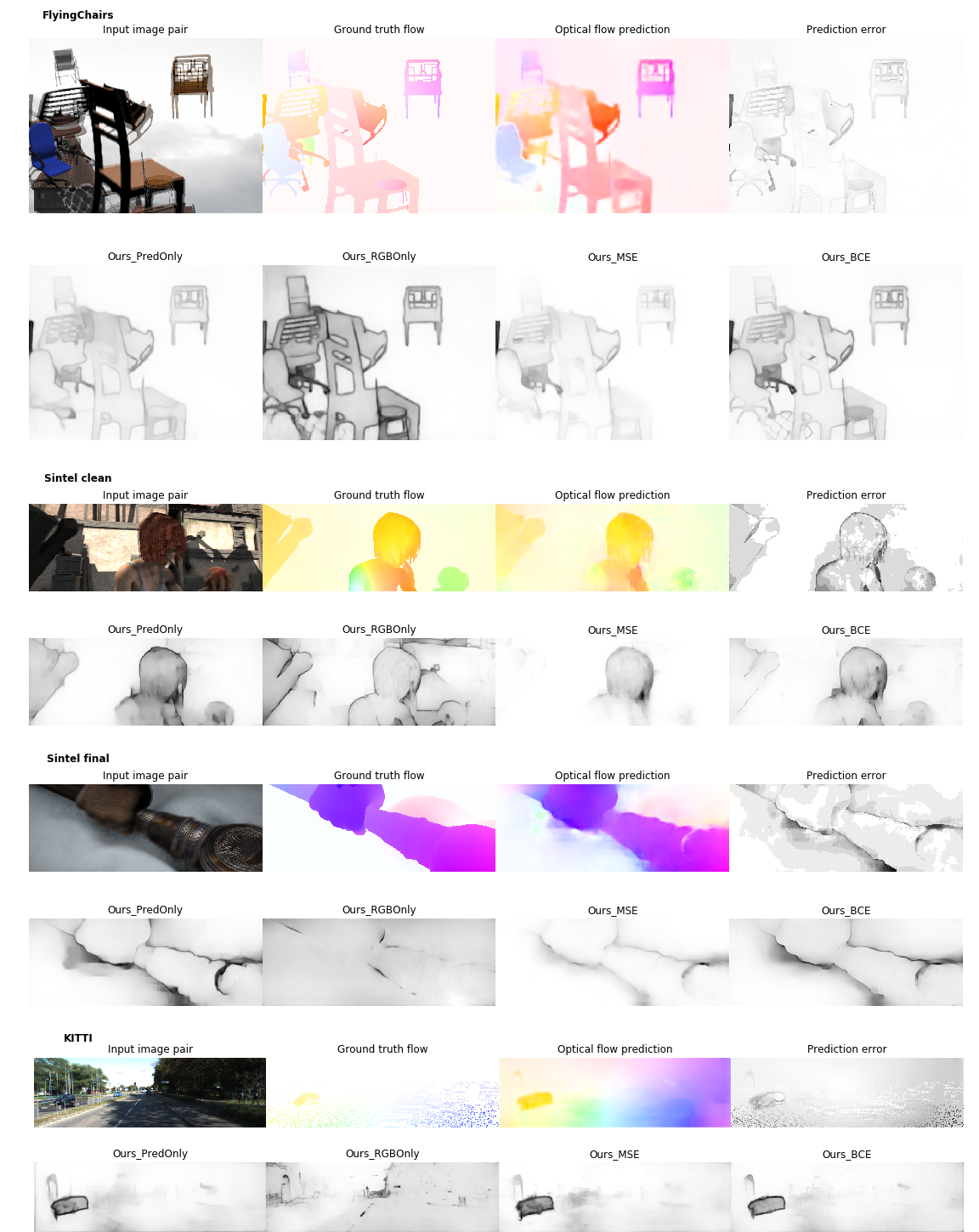}}
    \caption{\small{Uncertainty estimation examples in ablation study for OF task. The first row of each dataset block represents the input image pair, ground truth and predicted optical flow and the prediction error. The prediction map and error map are made by a single FlowNetS model as an example. The second row of each dataset block represents the uncertainty results in using SLURP side learner with different inputs and different loss functions. Black indicates higher uncertainty, white indicates lower uncertainty.}}
    \label{fig:vis_of}
\end{figure}

\begin{figure}[t!]
    \centering{\includegraphics[width=1\linewidth]{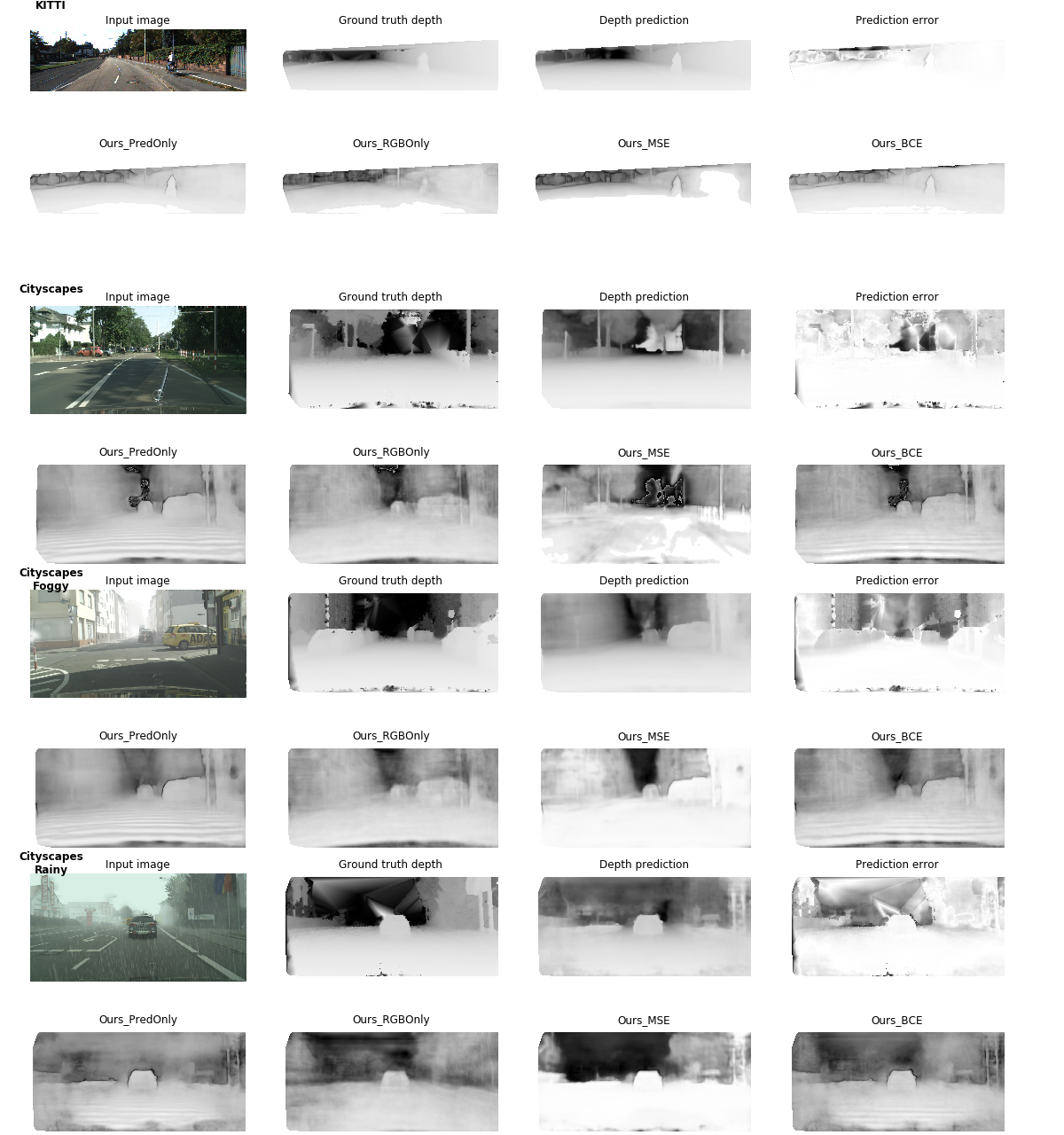}}
    \caption{\small{Uncertainty estimation examples in ablation study for MD task. The first row of each dataset block represents the input image, ground truth and predicted depth map and the prediction error. Since the ground truth is sparse, we use interpolation to rebuild the ground truth map just for visualization. The predicted depth map and error map are made by a single BTS model as an example. The second row of each dataset block represent the uncertainty results in using SLURP side learner with different inputs and different loss functions. For uncertainty maps, black indicates higher uncertainty, white indicates lower uncertainty. For depth maps, black represents deeper depth, and white represents shallower depth.}}
    \label{fig:vis_mde}
\end{figure}

For different inputs, we found that it is essential to use the prediction map as the input through the evaluation results. The input of the RGB image sometimes affects the generalization ability of uncertainty estimation if the main task model can generate already very good prediction results. According to the visualizations and evaluation results, we can see that the influence of the input of the prediction map is dominant because the uncertainty map of dual input and the one with only the prediction map input are similar. On the other hand, the RGB image can supplement some missing semantics of the prediction map, such as the Fig~\ref{fig:vis_of} FlyingChairs where RGB input can supplement the lack of chair legs and in the Fig~\ref{fig:vis_mde} where RGB input supplement the uncertainty of the sky (although the sky does not have ground truth of depth, it should have a high degree of uncertainty).
\clearpage
\begin{table}
\small
\begin{center}\resizebox{0.8\columnwidth}{!}{
\begin{tabular}{cccccc}
\hline
\multicolumn{1}{l}{\textbf{Conditions}} & \multicolumn{1}{l}{} & \multicolumn{1}{l}{} & \multicolumn{1}{l}{} & \multicolumn{1}{l}{} & \multicolumn{1}{l}{} \\ 
\hline
\multicolumn{1}{l}{\multirow{2}{*}{Input source}} & \multicolumn{1}{l}{RGB Input} & \checkmark & \xmark & \checkmark & \checkmark \\
\multicolumn{1}{l}{} & \multicolumn{1}{l}{Prediction map Input} & \xmark & \checkmark & \checkmark & \checkmark \\
\multicolumn{1}{l}{\multirow{2}{*}{Loss}} & \multicolumn{1}{l}{MSE} & \xmark & \xmark & \checkmark & \xmark \\
\multicolumn{1}{l}{} & \multicolumn{1}{l}{BCE} & \checkmark & \checkmark & \xmark & \checkmark \\ 
\hline
Datasets & Criteria & Ours RGBOnly & Ours PredOnly & Ours MSE & Ours BCE \\ 
\hline
\multirow{2}{*}{FlyingChairs} & AUSE-EPE & 1.82 & \textcolor{blue}{1.24} & 1.41 & \textbf{\textbf{1.20}} \\
 & AUROC & 0.944 & \textcolor{blue}{0.972} & 0.967 & \textbf{\textbf{0.974}} \\ 
\hline
\multirow{2}{*}{KITTI} & AUSE-EPE & 8.40 & \textcolor{blue}{4.87} & 5.40 & \textbf{\textbf{4.69}} \\
 & AUROC & 0.586 & \textbf{0.800} & \textcolor{blue}{0.793} & \textbf{\textbf{0.800}} \\ 
\hline
\multirow{2}{*}{Sintel Clean} & AUSE-EPE & 7.43 & \textbf{2.73} & 3.19 & \textcolor{blue}{2.91} \\
 & AUROC & 0.639 & \textbf{0.898} & 0.883 & \textcolor{blue}{0.896} \\ 
\hline
\multirow{2}{*}{Sintel Final} & AUSE-EPE & 8.24 & \textbf{2.71} & 3.11 & \textcolor{blue}{2.86} \\
 & AUROC & 0.575 & \textbf{0.907} & 0.889 & \textcolor{blue}{0.906} \\ 
\hline
 &  &  &  &  & 
\end{tabular}
}
\caption{\small{Ablation study for the OF task. Bold value: result with the best performance. Blue value: second performance.}}
\label{tab:ablation_of}
\end{center}
\end{table}

\begin{table}
\small
\begin{center}\resizebox{0.8\columnwidth}{!}{
\begin{tabular}{cccccc} 
\hline
\multicolumn{1}{l}{\textbf{Conditions}} & \multicolumn{1}{l}{} &  &  &  &  \\ 
\hline
\multicolumn{1}{l}{\multirow{2}{*}{Input source}} & \multicolumn{1}{l}{RGB Input~ ~~} & \checkmark & \xmark & \checkmark & \checkmark \\
\multicolumn{1}{l}{} & \multicolumn{1}{l}{Predinction map Input} & \xmark & \checkmark & \checkmark & \checkmark \\
\multicolumn{1}{l}{\multirow{2}{*}{Loss}} & \multicolumn{1}{l}{MSE~ ~ ~} & \xmark & \xmark & \checkmark & \xmark \\
\multicolumn{1}{l}{} & \multicolumn{1}{l}{BCE} & \checkmark & \checkmark & \xmark & \checkmark \\ 
\hline
Dataset & Criteria & Ours RGBOnly & Ours PredOnly & Ours MSE & Ours BCE \\ 
\hline
\multirow{3}{*}{KITTI} & AUSE-RMSE & 1.84 & 1.76 & \textcolor{blue}{1.74} & \textbf{\textbf{1.68}} \\
 & AUSE-Absrel & 4.45 & \textcolor{blue}{4.31} & \textbf{\textbf{4.19}} & 4.36 \\
 & AUROC & 0.879 & 0.890 & \textcolor{blue}{0.894} & \textbf{\textbf{0.895}} \\ 
\hline
\multirow{3}{*}{Cityscapes} & AUSE-RMSE & 9.95 & \textbf{9.40} & 9.82 & \textcolor{blue}{9.48} \\
 & AUSE-Absrel & 10.68 & \textbf{9.23} & \textcolor{blue}{10.29} & 10.90 \\
 & AUROC & 0.344 & \textbf{0.446} & \textcolor{blue}{0.414} & 0.400 \\ 
\hline
\multicolumn{6}{c}{After fine-tuning on Cityscapes} \\ 
\hline
\multirow{3}{*}{Cityscapes} & AUSE-RMSE & 3.47 & \textcolor{blue}{3.45 } & 4.77 & \textbf{\textbf{3.05}} \\
 & AUSE-Absrel & 6.71 & \textbf{6.47 } & 6.93 & \textcolor{blue}{6.55} \\
 & AUROC & 0.837 & \textcolor{blue}{0.844 } & 0.766 & \textbf{\textbf{0.849}} \\ 
\hline
\multirow{3}{*}{\begin{tabular}[c]{@{}c@{}}Cityscapes\\Rainy s=1\end{tabular}} & AUSE-RMSE & 3.97 & \textcolor{blue}{3.43 } & 4.80 & \textbf{\textbf{3.39}} \\
 & AUSE-Absrel & 6.95 & \textbf{5.52 } & 7.30 & \textcolor{blue}{5.62} \\
 & AUROC & 0.739 & \textbf{0.795 } & 0.68 & \textcolor{blue}{0.788} \\ 
\hline
\multirow{3}{*}{\begin{tabular}[c]{@{}c@{}}Cityscapes\\Rainy s=2\end{tabular}} & AUSE-RMSE & 3.98 & \textcolor{blue}{3.39 } & 4.92 & \textbf{\textbf{3.36}} \\
 & AUSE-Absrel & 6.68 & \textbf{5.16 } & 7.09 & \textcolor{blue}{5.28} \\
 & AUROC & 0.747 & \textbf{0.801 } & 0.689 & \textcolor{blue}{0.794} \\ 
\hline
\multirow{3}{*}{\begin{tabular}[c]{@{}c@{}}Cityscapes\\Rainy s=3\end{tabular}} & AUSE-RMSE & \textcolor{blue}{4.11} & \textbf{3.41 } & 5.07 & \textbf{\textbf{3.41}} \\
 & AUSE-Absrel & 6.77 & \textbf{4.85 } & 7.06 & \textcolor{blue}{5.05} \\
 & AUROC & 0.748 & \textbf{0.811 } & 0.694 & \textcolor{blue}{0.801} \\ 
\hline
\multirow{3}{*}{\begin{tabular}[c]{@{}c@{}}Cityscapes\\Foggy s=1\end{tabular}} & AUSE-RMSE & 3.55 & \textcolor{blue}{3.42 } & 4.92 & \textbf{\textbf{3.04}} \\
 & AUSE-Absrel & 6.40 & \textbf{6.15 } & 6.92 & \textcolor{blue}{6.25} \\
 & AUROC & 0.835 & \textcolor{blue}{0.841 } & 0.763 & \textbf{\textbf{0.847}} \\ 
\hline
\multirow{3}{*}{\begin{tabular}[c]{@{}c@{}}Cityscapes\\Foggy s=2\end{tabular}} & AUSE-RMSE & 3.51 & \textcolor{blue}{3.39 } & 5.03 & \textbf{\textbf{3.01}} \\
 & AUSE-Absrel & 6.23 & \textbf{5.98 } & 6.89 & \textcolor{blue}{6.06} \\
 & AUROC & 0.838 & \textcolor{blue}{0.845 } & 0.767 & \textbf{\textbf{0.852}} \\ 
\hline
\multirow{3}{*}{\begin{tabular}[c]{@{}c@{}}Cityscapes\\Foggy s=3\end{tabular}} & AUSE-RMSE & 3.48 & \textcolor{blue}{3.36 } & 5.24 & \textbf{\textbf{3.08}} \\
 & AUSE-Absrel & 5.97 & \textbf{5.72 } & 6.75 & \textcolor{blue}{5.80} \\
 & AUROC & 0.845 & \textcolor{blue}{0.852 } & 0.773 & \textbf{\textbf{0.857}} \\ 
\hline
 &  &  &  &  & 
\end{tabular}
}
\caption{\small{Ablation study for the MD task. Bold value: result with the best performance. Blue value: second performance. s (\emph{e.g} s=1) indicates severity, higher the s value, higher the severity.}}
\label{tab:ablation_MDE}
\end{center}
\end{table}


\clearpage

\section{Experiments}
\label{sec:expe}
\subsection{Evaluation protocol details}
\subsubsection{Optical flow}
Let us consider an optical flow dataset $D=\{(\mathbf{x}_i, \mathbf{y}_i)\}_i$, where  $\mathbf{y}_i\in{\mathcal{R}^2}$, $\mathbf{y}_i=(u_i, v_i)$ is the ground truth optical flow for pixel $\mathbf{x}_i$. Below, $\hat{\mathbf{y}_i}=(\hat{u}_i, \hat{v}_i)$ represents the optical flow prediction. \\
\textbf{1. End point error (EPE)}: The average end point error for valid pixels in $D$ is the Euclidean distance between $\mathbf{y}_i$ and $\hat{\mathbf{y}_i}$:
\begin{equation}
    EPE = \frac{1}{|D|}\sum_{(u_i,v_i)\in{D}}\sqrt{(u_i - \hat{u_i})^2 + (v_i - \hat{v_i})^2}
\end{equation}
EPE metric is used for illustrating the performance of the models we use in OF tasks shown in Table~\ref{tab:of_models} and it's also used in AUSE and AUROC evaluation for OF task.

\subsubsection{Monocular depth}
Let us consider a monocular depth dataset $D=\{(\mathbf{x}_i, d_i)\}_i$ where $d_i\in{\mathcal{R}^+}$ is the ground truth depth value for pixel $\mathbf{x}_i$. Below, $\hat{{d}_i}$ represents the depth prediction. The metrics we used in the evaluations are as follows:\\
\textbf{1. Root mean square error (RMSE)}:
\begin{equation}
    RMSE = \sqrt{\frac{1}{|D|}\sum_{d_i\in{|D|}}||\hat{d_i} - d_i||^2}
\end{equation}
\textbf{2. Absolute relative error (Absrel)}:
\begin{equation}
    Absrel = \frac{1}{|D|}\sum_{d_i\in{D}}|\hat{d_i} - d_i|/d_i
\end{equation}
\textbf{3. Threshold dk}: Inlier metrics as proposed in~\cite{eigen2014depth}, $k$ in dk indicates the power of the threshold ($thr$), we take $thr = 1.25$. In this case, d1: $thr = 1.25$; d2: $thr = 1.25^2$; d3: $thr = 1.25^3$, and dk represents the proportion of pixels that meet the threshold condition:
\begin{equation}
    dk = max(\frac{\hat{d_i}}{d_i}, \frac{d_i}{\hat{d_i}}) = \delta < thr^k
\end{equation}
\begin{equation}
    dk = \frac{|A|}{|D|}, \text{where } A = \left \{ \mathbf{x}_i, \text{such that } \delta_i = \max(\frac{\hat{d_i}}{d_i}, \frac{d_i}{\hat{d_i}}) \text{ and } \delta_i < thr^k \right \}
\end{equation}
\textbf{4. Squared relative difference (SqRel)}:
\begin{equation}
    SqRel = \frac{1}{|D|}\sum_{d_i\in{D}}{||\hat{d_i} - d_i||^2/d_i}
\end{equation}
\textbf{5. Root mean square log error (RMSElog)}:
\begin{equation}
    RMSElog = \sqrt{\frac{1}{|D|}\sum_{d_i\in{|D|}}{||\log\hat{d_i} - \log d_i||^2}}
\end{equation}
\textbf{6. Average log10 error (log10)}:
\begin{equation}
    log10 = \frac{1}{|D|}\sum_{d_i\in{|D|}}{|\log_{10}\hat{d_i} - \log_{10}d_i|}
\end{equation}
These six metrics measure the performance of the MD models we use, and are shown in Table~\ref{tab:depth_models}. At the same time, the first three metrics are also applied for AUSE and AUROC evaluations. Additionally, for AUROC, we choose $k=1$ for threshold dk metric.

\subsubsection{Sparsification plot settings}
For both MD and OF tasks, the area under the sparsification error curve (AUSE) made by the sparsification curve is computed image-wise and not dataset-wise in our evaluations because of the high memory consumption which would be required for sorting vaues across the entire dataset.\\
\textbf{Image-wise}: We calculate the standardized AUSE for every image in the test set by ranking its pixels according to the corresponding predicted uncertainty and true error, then calculate the average AUSE after traversing the entire test set.\\
\textbf{Dataset-wise}: By collecting all pixels of all images in the test dataset, we calculate the AUSE by sorting their predicted uncertainty and true error.

\subsection{Monocular depth estimation task supplement}\label{ssec:MDE_settings}
\subsubsection{Model precision}
Table~\ref{tab:depth_models} shows the main task model performance for different uncertainty estimation approaches~\cite{lakshminarayanan2016simple, kendall, gal2016dropout}. We have noticed that after the model is trained on KITTI, it cannot obtain reasonable accuracy on Cityscapes. This is because the ground truth of KITTI dataset is sparse, and only the lower half of the content is present. At the same time, the scene of Cityscapes is more complicated. Therefore, we fine-tune all models on Cityscapes to obtain reasonable accuracy.
\begin{table}[t!]
\small
\begin{center}\resizebox{\columnwidth}{!}{
\begin{tabular}{cc|ccc|ccccc} 
\hline
\multirow{2}{*}{Models} & \multirow{2}{*}{Datasets} & \multicolumn{3}{c|}{higher is better} & \multicolumn{5}{c}{lower is better} \\
 &  & d1 & d2 & d3 & Abs Rel & Sq Rel & RMSE & RMSE log & log10 \\ 
\hline
\multirow{2}{*}{MC} & KITTI & 0.945 & 0.992 & 0.998 & 0.072 & 0.287 & 2.902 & 0.107 & 0.031 \\
 & Cityscapes & 0.103 & 0.255 & 0.453 & 1.051 & 18.942 & 18.986 & 0.842 & 0.324 \\ 
\hline
\multirow{2}{*}{EE (DE)} & KITTI & 0.957 & 0.993 & 0.999 & 0.059 & 0.233 & 2.688 & 0.093 & 0.026 \\
 & Cityscapes & 0.214 & 0.430 & 0.560 & 0.837 & 14.459 & 18.441 & 0.845 & 0.298 \\ 
\hline
{Ours~} & KITTI & 0.955 & 0.993 & 0.998 & 0.060 & 0.249 & 2.798 & 0.096 & 0.027 \\
(Single-PU, Original~\cite{lee2019big}) & Cityscapes & 0.183 & 0.386 & 0.519 & 0.963 & 17.230 & 18.948 & 0.896 & 0.321 \\ 
\hline
\multicolumn{10}{c}{After fine-tuning on Cityscapes} \\ 
\hline
MC & Cityscapes & 0.882 & 0.974 & 0.992 & 0.117 & 0.917 & 5.625 & 0.169 & 0.049 \\ 
\hline
EE (DE) & Cityscapes & 0.920 & 0.983 & 0.995 & 0.098 & 0.635 & 4.889 & 0.149 & 0.043 \\ 
\hline
Ours (Single-PU) & Cityscapes & 0.906 & 0.980 & 0.993 & 0.104 & 0.711 & 5.216 & 0.159 & 0.046 \\ 
\hline
 & \multicolumn{1}{c}{} &  &  & \multicolumn{1}{c}{} &  &  &  &  & 
\end{tabular}
}
\caption{\small{The performance on KITTI and the performance on Cityscapes dataset before and after fine-tuning the models. Before fine-tuning: Training set: KITTI Eigen-split training set, Test set: KITTI Eigen-split test set and Cityscapes test set; After fine-tuning: Fine-tuning training set: Cityscapes training set, Test set: Cityscapes test set. \Xuanlong{The three main task models used in EE are also used in DE and Single-PU also shares the same main task model with our approach.}}}
\label{tab:depth_models}
\end{center}
\end{table}

\subsubsection{Training settings}
In the MD task, we choose to use a sequential training strategy for single predicted uncertainty (Single-PU)~\cite{kendall}, deep ensembles (DE)~\cite{lakshminarayanan2016simple} and our SLURP side learner. In other words, we first complete the training of the main task models (BTS~\cite{lee2019big}) and then train different uncertainty predictors according to the settings. Specifically, for Single-PU, we use an identical BTS model to estimate the uncertainty with using the output of the main task its corresponding main task model in the loss. For DE, it is a mixture of multiple Single-PUs, so we just repeat the previous procedures. Because of the ensemble property of empirical ensemble (EE) and DE, EE and DE can share the same main task predictors. In the same sense, the main task predictor of our side learner is chosen from one of EE(DE)'s main task predictors, which is the same one as the main task model of Single-PU. This method can ensure that the prediction accuracy of the main task will not be affected by the training of the uncertainty predictor. The ConfidNet~\cite{corbiere2019addressing} (Confid) implementation for BTS references its implementation on SegNet. The detailed operations are consistent with the descriptions in the main paper. \\
We build our side learner according to SLURP solution in the main paper (also shown in Fig 2 in the main paper) for BTS and here are some supplements. We directly use the frozen RGB feature maps from the encoder of main task BTS model. To convert 1-channel predicted depth map to 3-channel input, we expand it three times. The detailed settings for different uncertainty estimation models are listed in Table~\ref{tab:mde_settings}, all main task models are trained identically according to the original BTS~\cite{lee2019big} model training settings.
\begin{table}[t!]
\small
\begin{center}\resizebox{1\columnwidth}{!}{

\begin{tabular}{l|cccc} 
\hline
\multicolumn{1}{c|}{Hyper-parameters} & MC & EE & DE (Single-PU, Confid) & Ours \\ 
\hline
\begin{tabular}[c]{@{}l@{}}learning rate for\\main task model (Training on KITTI)\end{tabular} & 1e-4 & 1e-4 & / & / \\ 
\hline
number of training epoch

~(Training on KITTI) & 50 & 50 & 50 & 8 \\ 
\hline
\multicolumn{1}{l}{learning rate forside learner 

~(Training on KITTI)} & / & / & / & 1e-4 \\ 
\hline
\begin{tabular}[c]{@{}l@{}}learning rate for\\main task model (Fine-tuning on Cityscapes)\end{tabular} & 5e-5 & 5e-5 & / & / \\ 
\hline
number of training epoch (Fine-tuning on Cityscapes) & 30 & 30 & 30 & 16 \\ 
\hline
\begin{tabular}[c]{@{}l@{}}learning rate for\\side learner (Fine-tuning on Cityscapes)\end{tabular} & / & / & / & 8e-5 \\ 
\hline
\begin{tabular}[c]{@{}l@{}}learning rate for identical uncertainty estimator\end{tabular} & / & / & 5e-5 & / \\ 
\hline
\begin{tabular}[c]{@{}l@{}}learning rate for side learner\end{tabular} & / & / & / & 1e-4 \\ 
\hline
\begin{tabular}[c]{@{}l@{}}batch size\end{tabular} & \multicolumn{4}{c}{4} \\ 
\hline
number of training epoch & 50 & 50 & 50 & 8 \\ 
\hline
\begin{tabular}[c]{@{}l@{}}weight decay for\\main task model\end{tabular} & 1e-2 & 1e-2 & / & / \\ 
\hline
\begin{tabular}[c]{@{}l@{}}weight decay for\\identical uncertainty estimator\end{tabular} & / & / & 1e-2 & / \\ 
\hline
\begin{tabular}[c]{@{}l@{}}weight decay for\\side learner feature extractor\end{tabular} & / & / & / & 1e-3 \\ 
\hline
\begin{tabular}[c]{@{}l@{}}weight decay for\\side learner uncertainty generation blocks\end{tabular} & / & / & / & 4e-4 \\ 
\hline
\multicolumn{5}{c}{Model structure and other settings~} \\ 
\hline
\begin{tabular}[c]{@{}l@{}}encoder backbone for main task model, \\identical uncertainty estimator and \\side learner feature extractor\end{tabular} & \multicolumn{4}{c}{Densenet 161~\cite{huang2017densely}} \\ 
\hline
loss & same as BTS~\cite{huang2017densely} & same as BTS~\cite{huang2017densely} & Laplacian NLL & \begin{tabular}[c]{@{}c@{}}BCE\\$\lambda = 0.0125$\end{tabular} \\ 
\hline
number of latent stages $n$ & / & / & / & 5 \\ 
\hline
number of latent stage output channel $c$ & / & / & / & 1 \\ 
\hline
number of final uncertainty output channel $C_{out}$ & / & / & / & 1 \\ 
\hline
dropout rate $p_d$ & 0.4 & / & / & / \\ 
\hline
ensemble size $M$ & 1 & 3 & 3 (1) & 1 \\ 
\hline
\begin{tabular}[c]{@{}l@{}}during inference time\\number of forward propagation\end{tabular} & 8 & 3 & 3 (1) & 1 \\
\hline
\multicolumn{1}{l}{} & \multicolumn{1}{l}{} & \multicolumn{1}{l}{} & \multicolumn{1}{l}{} & \multicolumn{1}{l}{}
\end{tabular}
}
\caption{\small{MD model settings for MC, EE, DE, Single-PU, Confid and Ours.}}
\label{tab:mde_settings}
\end{center}
\end{table}

\subsection{Optical flow supplement}
\label{ssec:of}
\subsubsection{Model precision}
Table~\ref{tab:of_models} shows the main task precision for different uncertainty estimation strategies. Our SLURP side learner picks one of the models from EE as our main task predictor. The main task models are trained only on FlyingChairs training set with official split and the KITTI dataset we choose for main task precision evaluation and also uncertainty estimation/evaluation is KITTI 2015 with occlusions. In the original FlowNetS paper~\cite{dosovitskiy2015flownet}, the precision evaluation is based on KITTI 2012~\cite{Geiger2012CVPR}.
\begin{table}[t!]
\small
\begin{center}\resizebox{0.7\columnwidth}{!}{
\begin{tabular}{c|cccccc} 
\hline
Datasets & EE & MC & Single-PU & DE & Ours & Orginal~\cite{dosovitskiy2015flownet} \\ 
\hline
FlyingChairs test & 1.79 & 3.71 & 2.04 & 1.93 & 1.96 & 2.71 \\
KITTI 2015 occ & 18.36 & 16.53 & 21.21 & 20.78 & 19.39 & / \\
KITTI 2012 noc & 6.77 & 19.02 & 8.34 & 8.40 & 7.65 & 8.26 \\
Sintel clean train & 5.10 & 6.31 & 5.12 & 5.00 & 5.20 & 4.50 \\
Sintel final train & 6.50 & 6.97 & 6.53 & 6.41 & 6.62 & 5.45 \\ 
\hline
\multicolumn{1}{c}{} &  &  &  &  &  & 
\end{tabular}
}
\caption{\small{The main task accuracy for the uncertainty estimators in OF task. The values present the end-point error (EPE). Training set: FlyingChairs training set~\cite{dosovitskiy2015flownet}, Test set: FlyingChairs test set, KITTI 2012 noc~\cite{Geiger2012CVPR} which was used in the original FlowNetS paper, KITTI 2015 occ~\cite{geiger2013vision, Menze2015ISA, Menze2018JPRS} and Sintel full training set~\cite{Butler:ECCV:2012}}}.
\label{tab:of_models}
\end{center}
\end{table}
\subsubsection{Training settings}
In the optical flow task, for EE, we directly train multiple main task prediction models FlowNetS~\cite{dosovitskiy2015flownet}, and our side learner selects one of the models as our main task predictor. For Single-PU, because FlowNetS is relatively simple, we directly modify the original model to output two values for each pixel, one representing the predicted value of the main task and the other the uncertainty value. For DE, we train multiple Single-PU models. For multi-hypothesis prediction network (MHP)~\cite{ilg2018uncertainty}, we modified FlowNetS so that it can output eight (number of hypothesis) pairs of main task - uncertainty results. Futhermore, we use another FlowNetS as the MergeNet. It should be noted that, the authors did not mention the structural information about MergeNet in the paper. We choose FlowNetS based on the use of model stacking in the article. We train MHP followed by the two-stage training schedule provided in the supplementary of this paper.\\
For our SLURP side learner, since the encoder in FlowNetS is designed for capturing the object movement for two images and the total uncertainty will reflect only the semantics from the first image, we use two DenseNet161 backbones~\cite{huang2017densely} as RGB and prediction map encoders respectively. We also used two DenseNet121 backbones for the lighter version. In order to transfer the 2-channel flow prediction to a 3-channel input, we just add one convolution layer to expend the channel number before the RGB feature extractor. All uncertainty model training settings are shown in Table~\ref{tab:of_settings}.

\begin{table}[h!]
\small
\begin{center}\resizebox{1\columnwidth}{!}{

\begin{tabular}{l|cccc}
\hline
\multicolumn{1}{c|}{Hyper-parameters} & MC & EE & DE (Single-PU, Confid) & Ours \\ 
\hline
learning rate for (modified) main task model & 1e-4 & 1e-4 & 1e-4 & / \\ 
\hline
learning rate for side learner & / & / & / & 1e-4 \\ 
\hline
batch size & \multicolumn{4}{c}{8} \\ 
\hline
number of training epoch & 216 & 216 & 216 & 30 \\ 
\hline
weight decay for (modified) main task model & 4e-4 & 4e-4 & 4e-4 & / \\ 
\hline
weight decay for side learner feature extractors & / & / & / & 1e-4 \\ 
\hline
weight decay for side learner uncertainty generation blocks & / & / & / & 4e-4 \\ 
\hline
\multicolumn{5}{c}{Model structure and other settings} \\ 
\hline
loss & \begin{tabular}[c]{@{}c@{}}same as\\ FlowNetS~\cite{dosovitskiy2015flownet}\end{tabular} & \begin{tabular}[c]{@{}c@{}}same as\\ FlowNetS~\cite{dosovitskiy2015flownet}\end{tabular} & Laplacian NLL & \begin{tabular}[c]{@{}c@{}}BCE\\$\lambda = 0.05$\end{tabular} \\ 
\hline
number of latent stages $n$ & / & / & / & 5 \\ 
\hline
number of latent stage output channel $c$ & / & / & / & 2 \\ 
\hline
number of final uncertainty output channel $C_{out}$ & / & / & / & 1 \\ 
\hline
dropout rate $p_d$ & 0.4 & / & / & / \\ 
\hline
ensemble size $M$ & 1 & 3 & 3 (1) & 1 \\
\hline
during inference time number of forward propagation & 8 & 3 & 3 (1) & 1 \\ 
\hline
\multicolumn{1}{l}{} & \multicolumn{1}{l}{} & \multicolumn{1}{l}{} & \multicolumn{1}{l}{} & \multicolumn{1}{l}{}
\end{tabular}
}
\caption{\small{OF model settings for MC, EE, DE, Single-PU, Confid, MHP and Ours. MHP training setting is followed by the same schedule provided by is original paper.}}
\label{tab:of_settings}
\end{center}
\end{table}

\subsection{Synthetic 1D regression task supplement}
\label{ssec:toy}
Because of the simplicity of the data, the main task predictor we use is a neural network composed of one hidden layer and 3000 neurons. In SLURP joint-training, we train our main task model and side learner at the same time without freezing any layers, and in SLURP sequential-training, we train our side learner while freezing the main task and using the latent values of it. For the side learner, following general SLURP solution, we use the same hidden layer as the prediction result feature extractor and three hidden layers with 128, 64, 16 neurons respectively as the context block in the uncertainty generation block, since we have only one stage, there is no fusion block in the end. The training details for all uncertainty estimation approaches are listed in Table~\ref{tab:1d_settings}.\\
%Although our uncertainty side learner is more complex than the main task model,
We can give an insight that SLURP strategy can also work on 1D-regression tasks. In addition, the structure of the SLURP side learner is variable, and other uncertainty estimation methods are limited to the structure of the main task model, we are able to get better uncertainty results.

\begin{table}[h!]
\small
\begin{center}\resizebox{\columnwidth}{!}{

\begin{tabular}{l|ccccc} 
\hline
\multicolumn{1}{c|}{Hyper-parameters} & MC & EE & DE (Single-PU) & \begin{tabular}[c]{@{}c@{}}SLURP\\joint-training\end{tabular} & \begin{tabular}[c]{@{}c@{}}SLURP\\sequential-training\end{tabular} \\ 
\hline
number of main task latent features & \multicolumn{5}{c}{3000} \\ 
\hline
learning rate for main task model & 1e-1 & 1e-1 & 1e-2 & 1e-1 & / \\ 
\hline
\begin{tabular}[c]{@{}l@{}}learning rate for\\side learner feature extractor\end{tabular} & / & / & / & 1e-1 & 1e-4 \\ 
\hline
\begin{tabular}[c]{@{}l@{}}learning rate for\\side learner uncertainty generation blocks\end{tabular} & / & / & / & 1e-4 & 1e-4 \\ 
\hline
batch size & \multicolumn{5}{c}{50} \\ 
\hline
number of training epoch & \multicolumn{5}{c}{50} \\ 
\hline
\begin{tabular}[c]{@{}l@{}}weight decay for\\main task model\end{tabular} & 1e-2 & 1e-2 & 1e-2 & 1e-2 & / \\ 
\hline
\begin{tabular}[c]{@{}l@{}}weight decay for\\side learner feature extractor\end{tabular} & / & / & / & 1e-2 & 1e-3 \\ 
\hline
\begin{tabular}[c]{@{}l@{}}weight decay for\\side learner uncertainty generation blocks\end{tabular} & / & / & / & 1e-3 & 1e-2 \\ 
\hline
\multicolumn{6}{c}{Model structure and other settings} \\ 
\hline
loss & \multicolumn{1}{l}{MSE} & \multicolumn{1}{l}{MSE} & \multicolumn{1}{l}{Gaussian NLL} & \multicolumn{1}{l}{Gaussian NLL} & \multicolumn{1}{l}{MSE} \\ 
\hline
dropout rate & 0.4 & / & / & / & / \\ 
\hline
ensemble size M & 1 & 3 & 3 (1) & 1 & 1 \\ 
\hline
\multicolumn{1}{l}{} & \multicolumn{1}{l}{} & \multicolumn{1}{l}{} & \multicolumn{1}{l}{} & \multicolumn{1}{l}{} & \multicolumn{1}{l}{}
\end{tabular}
}
\caption{\small{1D regression task model settings.}}
\label{tab:1d_settings}
\end{center}
\end{table}

\clearpage

\section{More visualization results}
\label{sec:moar}
\begin{figure}[h!]
    \centering{\includegraphics[width=1\linewidth]{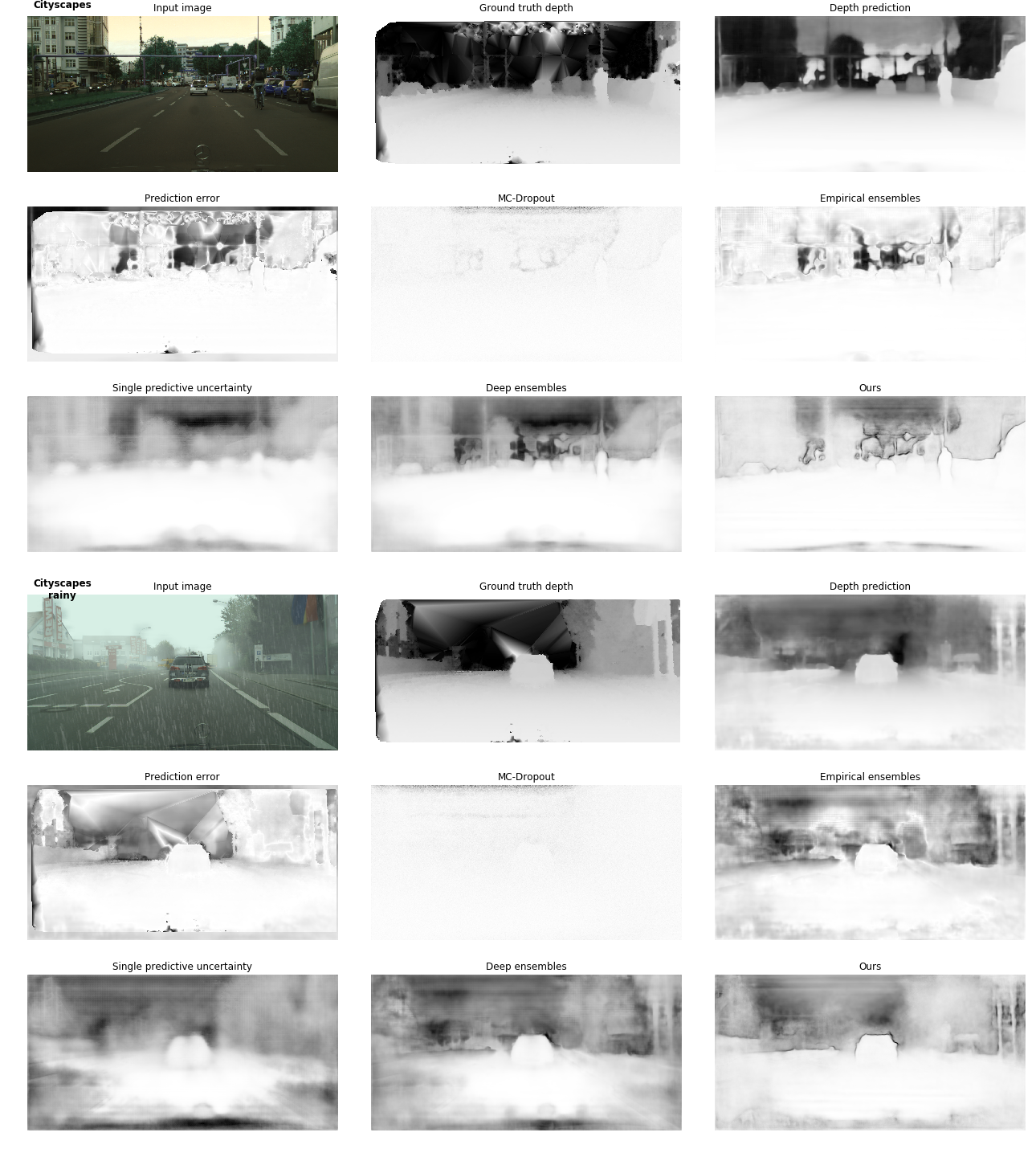}}
    \caption{\small{Uncertainty estimation results for MD task. The ground truth maps are rebuilt by interpolation just for visualization. The depth prediction map and the error map are generated by a single BTS model as an example. MC-Dropout uncertainty maps are obtained by eight forward propagation, Deep ensembles and Empirical ensembles uncertainty maps are obtained from three models ensembles. For uncertainty maps, black indicates higher uncertainty, white indicates lower uncertainty. For depth maps, black represents deeper depth, and white represents shallower depth.}}
    \label{fig:mde_vis1}
\end{figure}
\begin{figure}[h!]
    \centering{\includegraphics[width=1.1\linewidth]{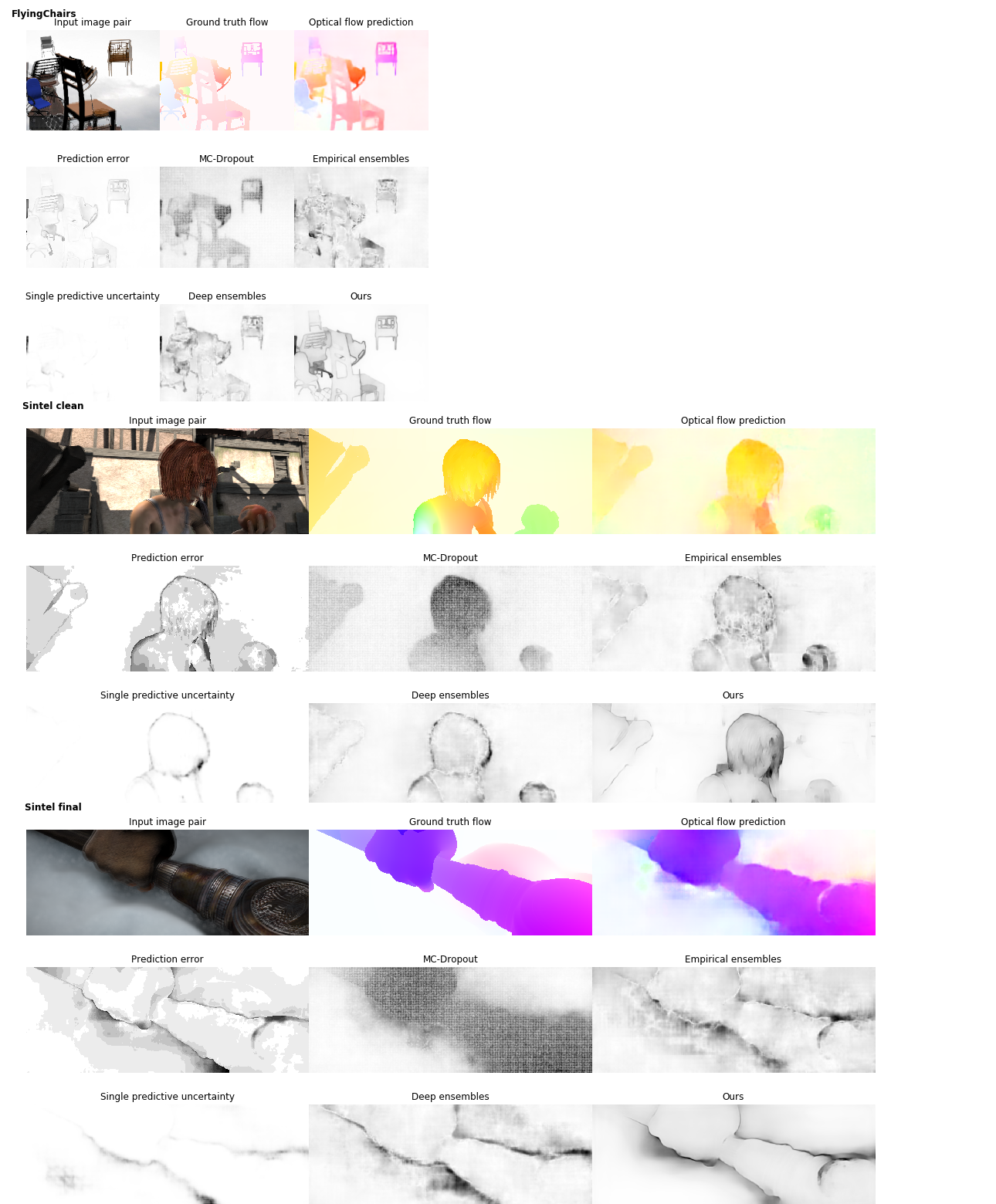}}
    \caption{\small{Uncertainty estimation results for OF task. The optical flow prediction map and the error map are generated by a single FlowNetS model as an example. MC-Dropout uncertainty maps are obtained by eight forward propagation, Deep ensembles and Empirical ensembles uncertainty maps are obtained from three models ensembles. For uncertainty maps, black indicates higher uncertainty, white indicates lower uncertainty.}}
    \label{fig:of_vis1}
\end{figure}
\clearpage